# Hardware and Software manual for Evolution of Oil Droplets in a Chemo-Robotic Platform

Juan Manuel Parrilla Gutierrez[1], Trevor Hinkley[1], James Taylor[1], Kliment Yanev[2], and Leroy Cronin[*1]

[1]WestCHEM, School of Chemistry, University of Glasgow, Glasgow, G12 8QQ UK

[2]Future Bits OpenTech UG, Cologne, 51103, Germany

[*]Corresponding author: lee.cronin@glasgow.ac.uk

## S.1 Hardware

The robotic system specified in this document, *DropBot*, was built on the foundations laid by the RepRap 3D printer project[1]. This project was chosen as a starting point due to its open-source philosophy, expansive documentation and large community. These reasons were considered advantageous as they are expected to contribute to the easy replication and adoption of the *DropBot* paradim and implementation in other laboratories, as compared to a proprietary product. Early designs of *DropBot* (data not shown) aimed to use a RepRap 3d printer directly, through the replacing of the thermoplastic extruder with a liquid handling system. After early prototyping phases, it was decided that this platform was unsuitable and so a new design was formulated, re-using modular components from the RepRap project rather than its monolithic design. This section provides an overview of the methodology applied to the design and implementation of *DropBot*: Subsection S.1.1 gives an introduction to the limitations encountered in the direct conversion of RepRap into a liquid-handling robot and the development of the final design used for this publication, S.1.3 gives an overview of the digital control systems used and S.1.2.3 describes the 3d-printed, servo actuated syringe design that was developed for this robotic platform.

### S.1.1 Robot Frame

The primary limitations encountered in early prototyping phases, which attempted the direct modification of a RepRap 3d printer into a liquid handling robot, were the limited working area and the trans-location of the target stage, in the Y axis, rather than the manipulation apparatus. The most common design for a RepRap 3d printer has a working area of 20x20cm, which was was unsuitably small for the target experiments. The limitations imposed by the size of the experimentation apparatus, prescribed a target stage of around 50x40cm; no extant printer design, which also met the other demands, was discovered that would fulfil this criterion, at the time of project initiation. The second factor, the movement of stage along the Y-axis, rendered the design unsuitable for a liquid handling robot, as it introduced extreme turbulence to the liquid-phase target chemistry; essentially, the stage was behaving similarly to a shaker plate. In addition, the design involved the majority of the Y-actuation mechanism being physically located underneath the stage, occupying space that would optimally be allocated to the inclusion of a camera, for the analysis of the experiments, with a clean field of view.

To correct these deficiencies, it was decided that the manipulation apparatus should be located on an overhead XY-axis, above a static, glass staging area, with a camera located on a separate XY-axis being located underneath this stage, viewing the experiments through the glass from the bottom. Whilst this restricted the robot to working with glass apparatus, it was decided that, since this would be the normal mode of operation in any case, this limitation would not impact the real-world performance.

### S.1.1.1 Mechanical design

Figure 1 shows the final design, used for all methods in this publication. The primary modules of the design are:

1. The staging area, on which the experimentation apparatus is placed.
2. The X-axis, which actuates motion along the long edge of the robot.
3. The Y-axis, which actuates motion along the short edge of the robot.
4. The manipulation apparatus (the mobile carriage), which performs formulation mixing, aqueous-phase handling, droplet placement and cleaning.
5. The fluid-handling platform, which handles the introduction of liquids to the main stage of the robot for further manipulation.

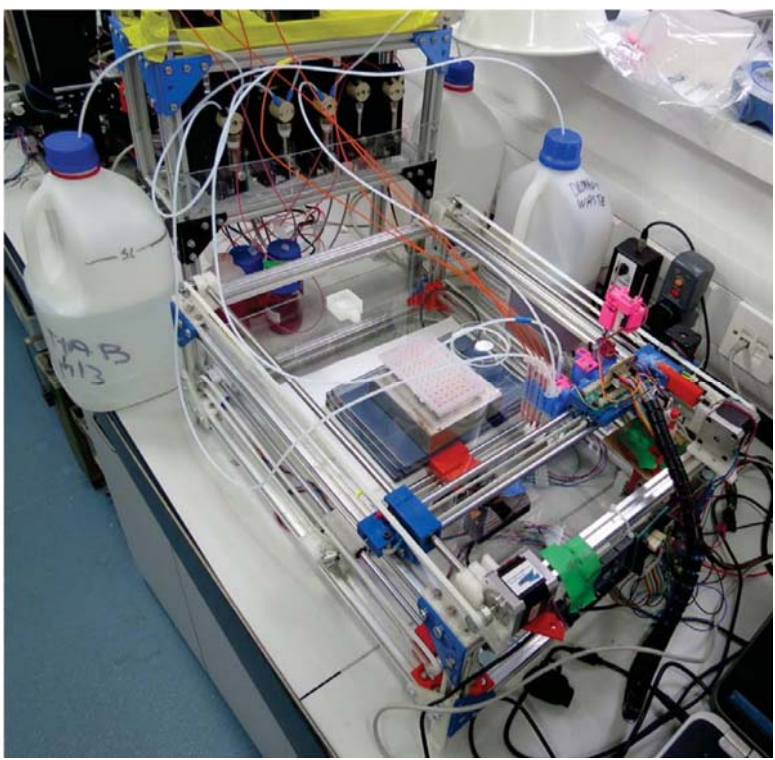

Figure 1: Overall picture of the robot. Strut profiles were used to build the frame. They were joined using 3D printed pieces. The carriage which holds the syringes and tubes was also 3D printed. All the mechanisms were based on RepRap 3D printers. The long axis shown here is the X axis, with the shorter being the Y axis. The carriage is shown at home (0,0) position.

### S.1.1.2 Staging area

The staging area consisted of a large glass plate, fixed to the robot frame. This area was defined by being that space over which the X-Y carriage could move and hence all apparatus that needed manipulation by the carriage was placed on the stage. At the centre of the stage was a 96-well plate, in which fluids were placed for formulation mixing, prior to droplet placement. There was a magnetic stirrer underneath the well-plate, to actuate miniature magnetic stirrer bars inside each well. Also atop the stage were two petri dishes, one was used to carry out the experiments and the other was used to collect waste after experiments were concluded. Manual intervention was required after a series of 48, or 96, experiments (dependant on whether experiments ran overnight) to clean the well-plate and waste petri dish.

### S.1.1.3  X-axis

Figure 2 shows a 3D-rendering of the X-axis translation mechanism. The same design was mirrored on both sides of the frame. This mechanism is static in relation to the staging area and frame of the robot.

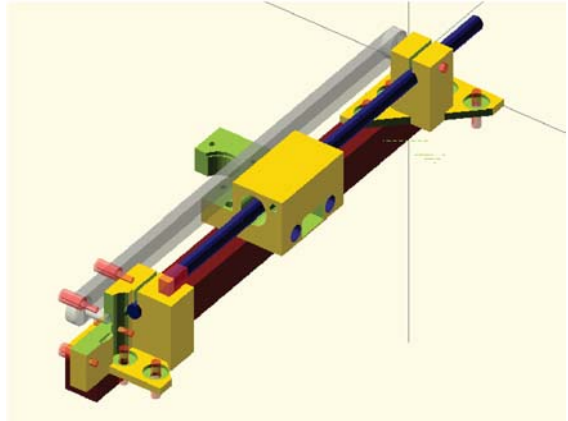

Figure 2: X axis 3D design. **Blue**: The precision rod (8h7). **Grey**: The timing belt (T2.5x6mm). **Yellow**: The 3d-printed parts (see section S.6). **Dark red**: Aluminium strut profile (20x20mm). **Transparent red**: The parallelepiped near the motor is where the end stops were placed; other parts in this colour are machine screws used to fasten the parts together.

### S.1.1.4  Y-axis

Figure 3 shows a 3d-rendering of the Y-axis translation mechanism. This mechanism runs along the X-axis via the X-axis belt, rod and linear bearing system. The two round steel bars seen in figure 3 were inserted into the complimentary holes seen in figure 2. A single motor was mounted on one of the X-axis carriages to actuate the central belt and provide linear motion along the Y-axis.

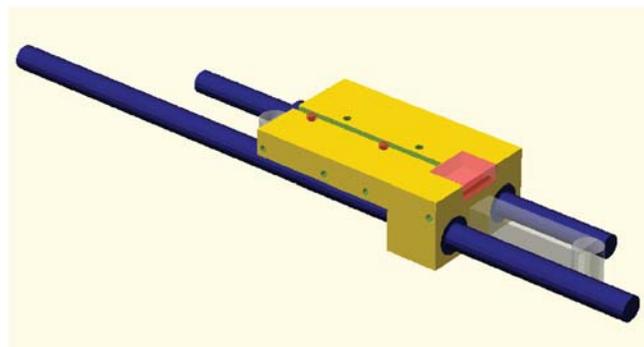

Figure 3: Y axis 3D design. **Blue**, the precision rod. **Grey**, the belt. **Yellow** the printed parts. The motor that moved the carriage around the Y axis could be seen on Figure 2. The red transparent parallelepiped is where the end stops were placed. The four holes on its side, and on the other side where it cannot be seen, were used to place the structures that would handle the syringes or any other equipment, making the design customizable.

### S.1.1.5 Mobile carriage

Figure 4 shows a 3d-rendering of the X-Y carriage. This components runs along the Y-axis via the Y-axis belt, rod and linear bearing system. The "ridge" running from left to right along the Y-axis runner served as an attachment point for the circuitry (subsection S.1.3). The design of the syringe actuation mechanism is detailed in subsection S.1.2.3.

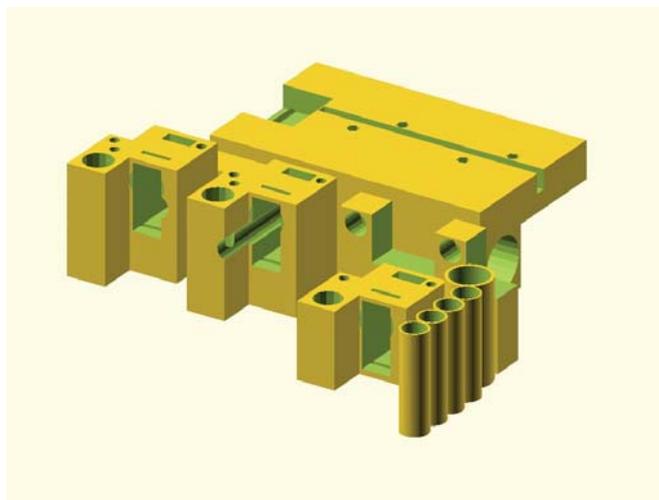

Figure 4: The X-Y carriage, assembled but without syringes (see subsection S.1.2.3). The locations for actuated syringes and fluidic connections to the fluid platform can be seen in the foreground. In the background is the attachment to the Y-axis.

### S.1.1.6 Fluid platform

The fluid-handling platform is a simple frame, constructed from aluminium strut profile and the 3d-printed assembly pieces used in the main robotic frame. Sheet-polycarbonate was attached to the strut-frame to support the weight of the pumps. The frame was designed so that reagents could be places underneath the pumps; This design was chosen to minimize the effect of any chemical spills by prevented contact with the electronics. The design of the pumps themselves is detailed in section S.1.2.1.

## S.1.2 Liquid Handling

Both servo-actuated syringes and syringe pumps were used to transport liquid phases. The servo-syringes were more precise but were unable to carry as much volume as the syringe pumps. Servo-syringes were used to mix the liquid in the well-plates and to carry small volumes of liquid into the petri dish via droplet formation. Syringe pumps were used to transport liquids from source reagent bottles to components on the experiment stage.

### S.1.2.1 Pumps

In the course of extraneous and prior research, the authors had accumulated a number of defective commercial syringe pumps, rendered inoperative through faults in their logic boards. Its original electronics were removed and the motors were connected to custom components (see subsection S.1.3). A total of seven such pumps were used by the robot, mounted on a single platform (see subsection S.1.1.6). Each pump was equipped with a threeway valve: Allowing the syringe to be connected through one port to either of the other two. One of these port was designated as an input port and the other as an output port. Three of the pumps were equipped with 5ml syringes: One was used for introducing the aqueous phase (see subsection S.3.1), and the other two for the introducing and removing solvents as part of the cleaning procedure (see subsection S.3.3). All pumps except one were connected via the input port to a reagent bottle and via the output port to the X-Y carriage; the sole except was the syringe pump used for removing the acetone from the petri dish as part of the cleaning cycle. The remaining four pumps were equipped with 1ml syringes; these were used to introduce organic phases, as described in subsection S.3.2.

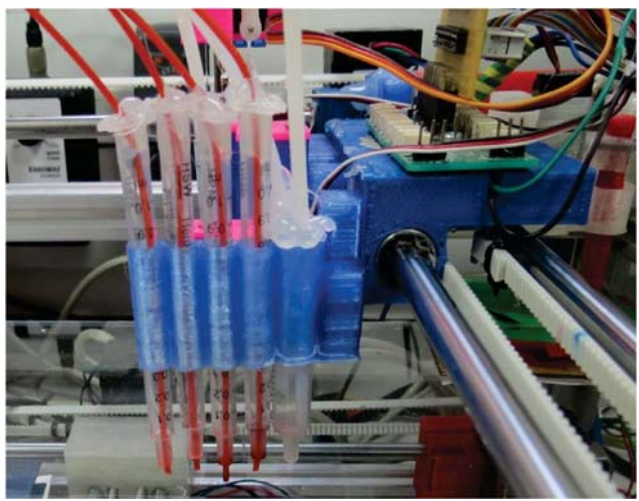

Figure 5: Tubing output setup. Plastic syringes barrels were used to guide the tubes. First the four oils, and at the end acetone and aqueous phase share the same barrel.

The tubing set-up at the X-Y carriage can be seen in figure 5. The syringe barrels were used as guides, to maintain accurate positioning of the tubing above the working area. Acetone and aqueous phase shared the same barrel, with two independent tubes being fixed inside by hot glue. The support structure was 3d-printed in PLA and was attached to the carriage with brass screws.

### S.1.2.2 Mixing stage

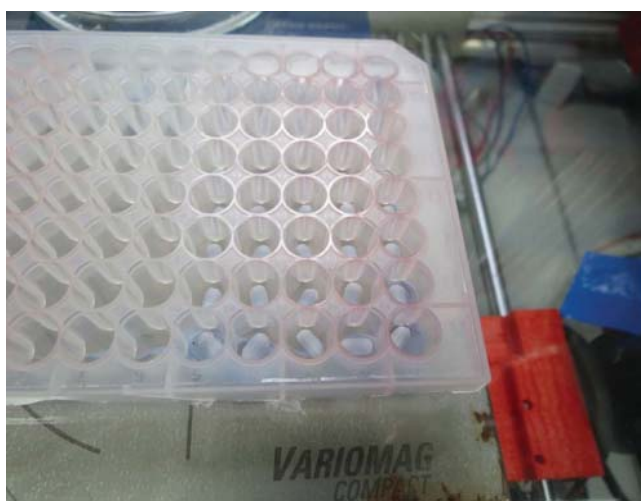

Figure 6: "Nunc U96 0.5 ml" well plates, with 'Magnetic stir bar micro PTFE 6 mm x 3 mm" and "Variomag Compact" stirrer plate used the mix the oils.

The robot used a standard 96-well plate (see figure 6); each well had its own miniature magnetic stirrer bar and the entire plate was mounted over a stirrer plate. This design was chosen so that multiple experiments could be concluded before manual intervention was necessary to clean the mixing area.

### S.1.2.3  Carriage-mounted syringes

As shown in figure 7, an automatic syringe was designed to be used with the X-Y carriage. The casing and structure was 3d-printed from PLA. Independent crank mechanisms were used to actuate the plunger of the syringe and to lower and raise the syringe. These consisted of the default servo motorarm and a 3d-printed piece, joined via steel pins. The crank was aligned with the centre of the syringe itself to avoid unwanted lateral torque. Smalldiameter steel rods and teflon linear bearings were used to mitigate rotation away from and towards the servo.

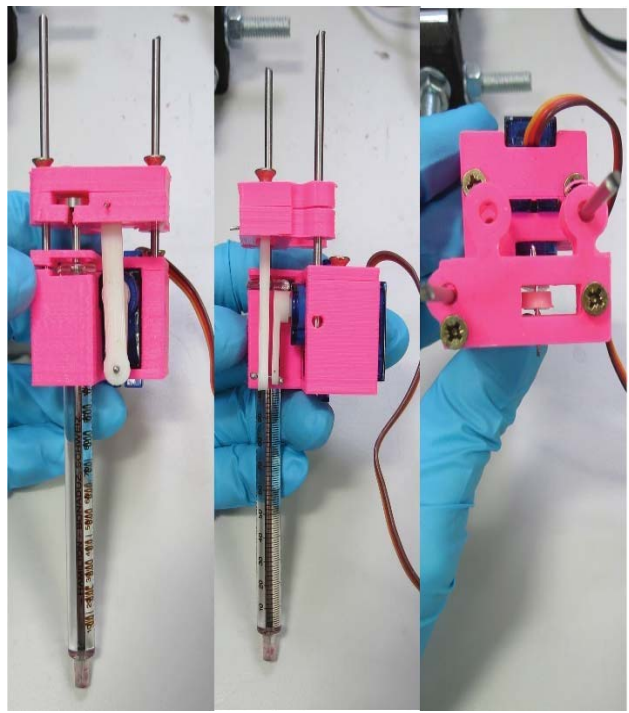

Figure 7: Automated syringe prototype. The plunger was actuated using a 9g servo motor. Guide rods were used to avoid bending.

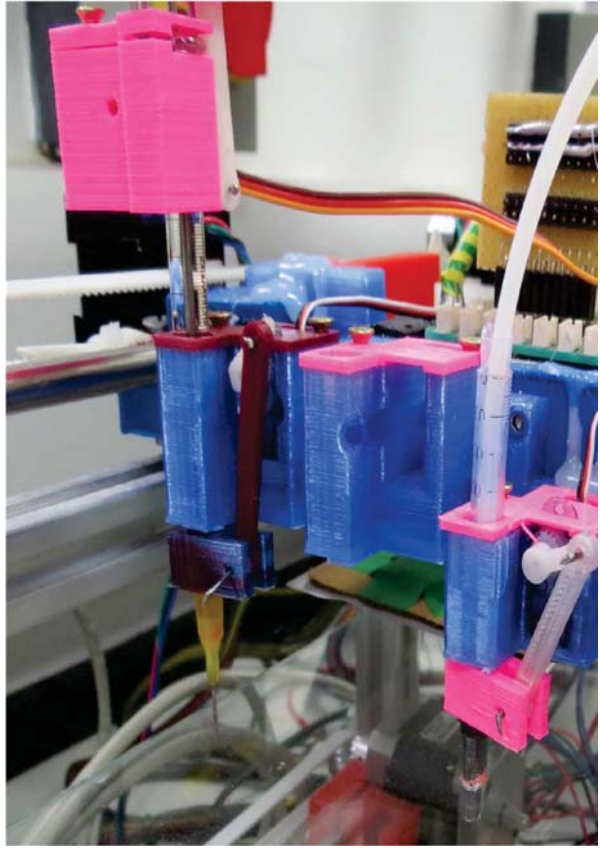

Figure 8: Syringe attached to needle and lifting mechanism. The same lifting mechanism was used to raise and lower a plastic needle to remove liquid from the Petri dish

The syringes were fitted with a metal needle-tip, as shown in figure 8. The mechanism to raise and lower the syringe, via a crank, can also be seen in this figure. On the right of figure 8, the same mechanism can be seen to be actuating an unactuated syringe. This syringe has had its plunger removed and is instead fitted with a piece of plastic tubing, which connects it to a syringe pumps. This syringe was used to remove solvents from the Petri dish during a cleaning cycle. For this reason, the needle taper tip was cut, to effect a larger absorption diameter. Whilst *DropBot* only used one of each syringe type, it would be possible to expand the robot by adding additional syringes to the carriage.

### S.1.3    Electronics

The electronics can be divided into three groups. One group controls the movement of the robot along the X and Y axes, another group controls the servo motors which actuate the syringes, and a third group control the pumps.

### S.1.3.1 Axis control

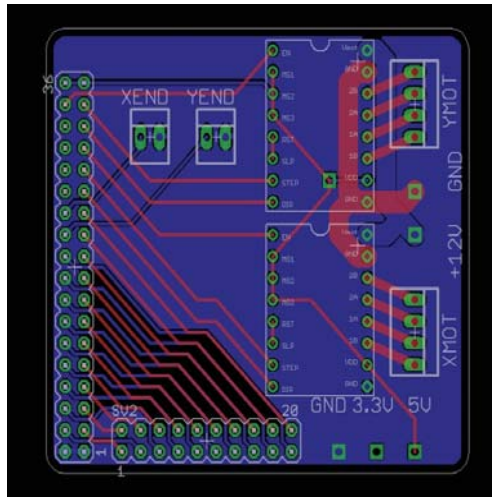

Figure 9: Arduino Mega PCB shield with direct control outputs to 2 stepper motors and breakout header to the servo daughter-board.

The movement along the X and Y axes was performed using a stripped down version of the electronics used by a RepRap 3D printer, where only the parts required by the X and Y movement were kept: an Arduino Mega[2] board and "A4988 Stepper Motor Driver Carrier". A PCB shield was designed (see Figure 9), which interfaced the Arduino board to the motor drivers and provided connectors for the motors to the drivers. In addition, connections were provided for two end-stops, each assigned to a particular axis, for homing purposes. Each stepper driver had 16 pins, but only 3 of them were needed to control the motor: enable, step and direction, with the other pins being connected to high (5V), low (GND), 12V or to the motor itself. The board therefore used only eight outputs (the end-stops used two outputs) from the Arduino directly and interfaced the remainder of digital connectors, and a power supply, to a header output for the servo daughterboard. The attachment between the Arduino board and the PCB shield can be seen on Figure 10.

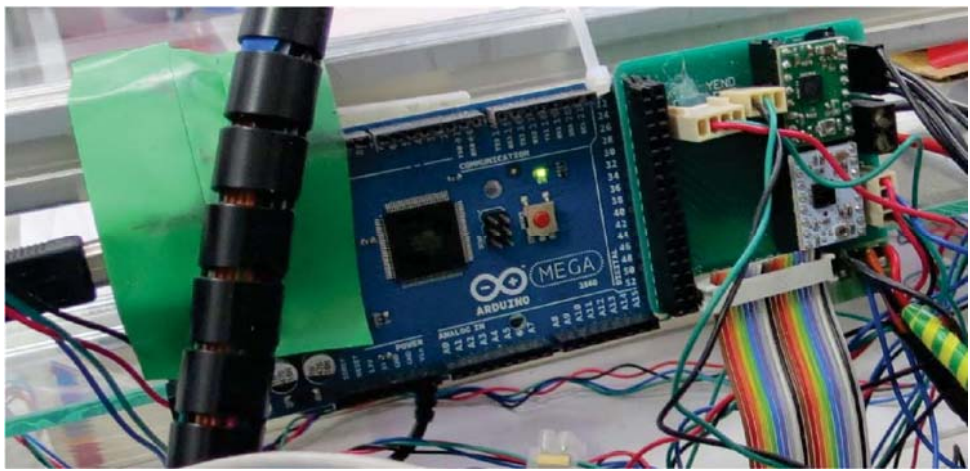

Figure 10: Arduino Mega board with the PCB shield described on Figure 9 attached.

### S.1.3.2  Carriage control

The header input to the servo daughter-board (Figure 11) consisted of twenty pins, connected to the Arduino via a ribbon cable to the bypass on the XY driver board. The servo board itself was mounted on the X-Y carriage in order to provide both power and control to the syringe-actuation servo motors. Each servo motor took power from a common supply and were connected to a common ground, with the only individual connection being a single PWM data pin to the Arduino. The servo daughter-board was therefore able to service 20 unique servo motors, of which maximum four were used in practice.

### S.1.3.3  Pump control

The pumps mentioned in section S.1.2.1 were faulty only in the function of their logic boards; the mechanical components and motors were fully oper-ational. As motion was provided by two stepper motors, from the NEMA family, these were therefore compatible with the shield developed for the X-Y axis control (Section S.1.3.1). It was therefore possible to re-use the mechanical components of these syringe pumps by the replacement of their electronics with custom PCBs (Figure 12). Since each board controlled two stepper motors, a single board was assigned to each pumps (Figure 13). A second Arduino was assigned entirely to pump control, in addition to a second power-supply assigned only to pump motors.

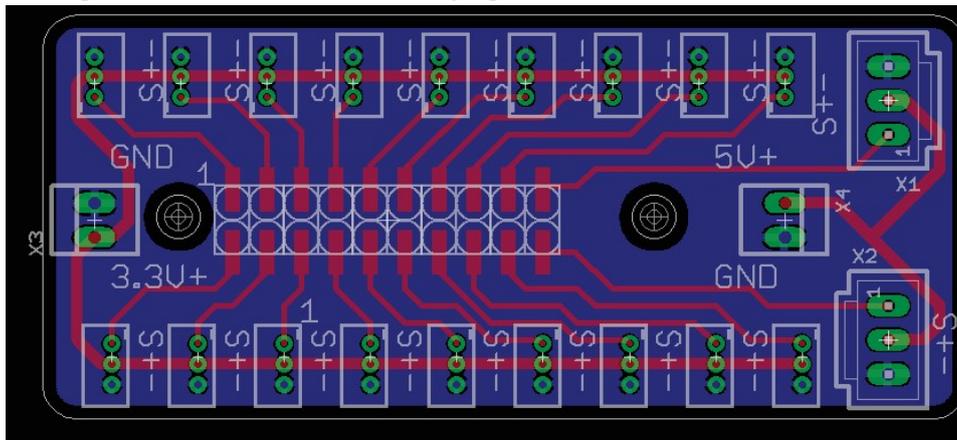

Figure 11: PCB designed to map the digital pins from the Arduino board into servo motor sockets. It worked for both 3.3V and 5V servo motors.

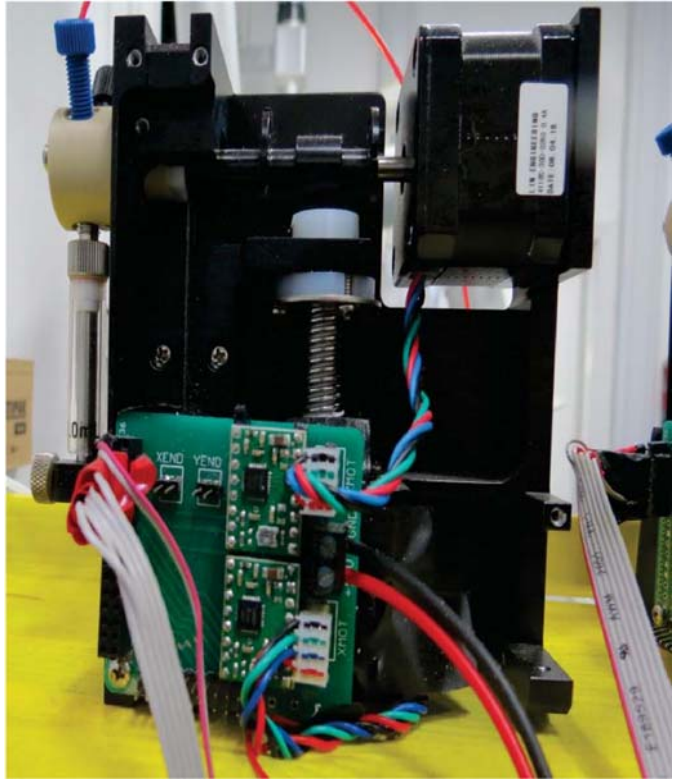

Figure 12: Tricontinent pump being used by attaching its motors to the designed PCB shield.

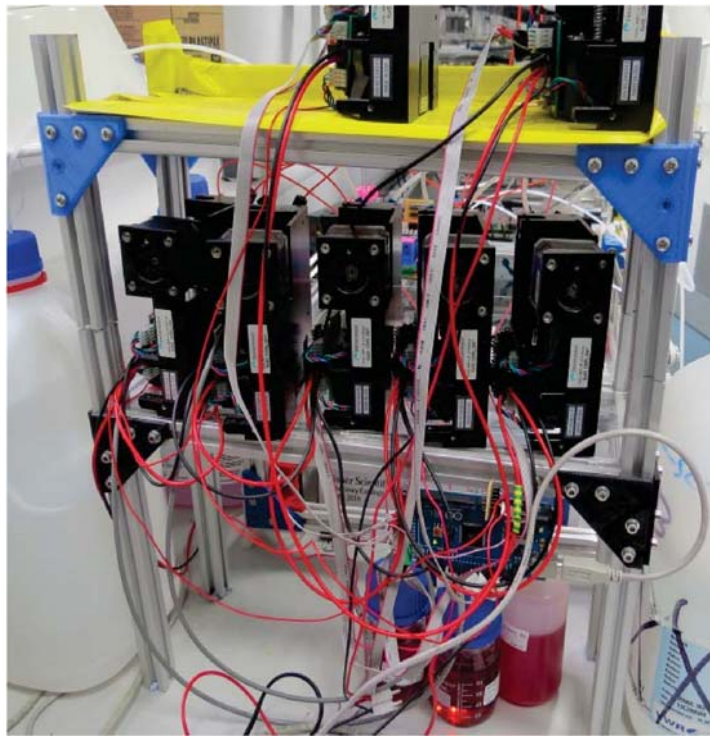

Figure 13: Each pump was assigned a unique PCB, with one driver controlling each motor. Each driver was controlled by an Arduino board common to all pumps but separate from the X-Y/servo Arduino.

### S.1.4 Bill of Materials

- The frame was built using Bosch-Wrexroth 20x20mm aluminium strut profile, fastened together using custom, 3d-printed PLA[1] pieces.

- The motors used were NEMA[2] 14 for the Y-axis and two NEMA 17s for the X-axis.

- The linear motion mechanics were derived from the RepRap printer, using a belt (Timing belt T2.5x6mm) and pulley (T2.5 pulley, 5mm bore) system connected to the motors

- Round-profile hardened-steel bar (8h7, chrome plated) and roundprofile linear bearings (LME8UU) were used to achieve smooth linear motion.

- The syringe pumps modified and used were "TriContinent C-Series" • "IDEX Health Science PEEK 1/8"" tubing was used to connect those pumps used introduce or remove cleaning solvent to/from the petri dish arena.

- "IDEX Health Science FEP Ora 1/16 x 0.20"" was used to carry organic phases from reagent bottles to syringe pumps and from syringe pumps to the X-Y Carriage.

- The syringe used to direct the tubing from the syringe pumps at the carriage the "1 ml NORM-JECT".

- These syringes were fitted with "Weller KDS16TN25 Needle Taper Tip 16G".

- The syringe used in the servo-actuated syringes was a "Hamilton 710 LT 100 $\mu L$". • The needles used were "Weller KDS2012P Dispensing Needle GA20 ID 0.66 MM".

- The syringe was raised and lowered by a "New Power XL-3.7" servomotor.

- The plunger was actuated with a "9g servo motor". In this specific case, a "TowerPro Micro Servo 9g SG90".

- The motor arm used was the default arm, which shipped with the servo motor.

---

[1] Polylactic acid.

[2] National Electrical Manufacters Association. http://www.nema.org

- The well plate used was "Nunc U96 PP 0.5 ml".

- Inside every well, a "Magnetic stir bar micro PTFE 6 mm x 3 mm" was used to provide liquid turbulence.

- Below the well plate, was a single "Variomag Compact" stirrer plate, used to turn the stirrer bars.

## S.2 Software

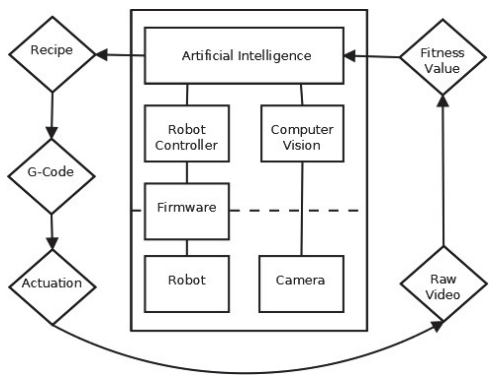

Figure 14: Software architecture showing distribution of workflow actors and actuators and their distribution into software packages.

A hierarchical/deliberative paradigm was followed when designing the control software. Software components resident on the host computer was programmed entirely in Python, whilst software components on the Arduinos (referred to henceforth as the firmware) were programmed in C++. Software was divided into three conceptual modules: *planning*, *acting*, and *sensing*. Figure 14 represents the structure of this hierarchy; data-flow occurs in an iterative loop (shown).

The artificial intelligence (AI) component is responsible for coordinating the overarching experimental plan, selecting droplet formulations based on prior (if any) data and passing these formulae on to the next component. Thus the AI acts as globally, with overview of an entire experimental series, and the RC, FW and CV components have purview only of single experiments.

The robotics controller (RC) takes these formulae as input and, acting as an interface between the AI and the physical robot, transforms these numerical recipes into a G-code representation. G-code is a standardized scheduling language[3] used in the automation industry to plan mechanical operations (see subsection S.2.2 for more details).

The firmware (FW) consists of software resident on the Arduino board. It converts the symbolic G-code representation into a series of experimental operations, instantiated as a sequence of analogue and digital signals sent to the mechanical parts through the Arduino boards.

The computer vision (CV) component interfaces with the camera and converts raw image data, through an algorithmic image-processing pipeline, into quantitative-numerical fitness data, which is then returned to the AI.

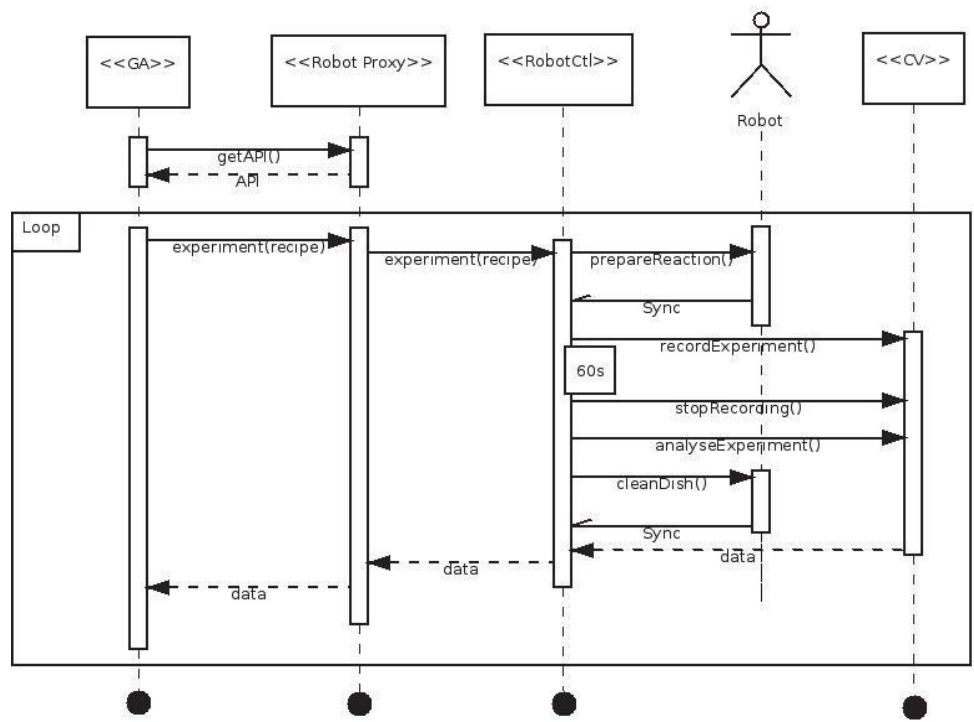

Figure 15: UML diagram describing the software work-flow and communication between the diverse software components.

A multiprocessing software architecture was chosen to make maximum usage of time and resources, and to ease development by modularizing the software. Software interprocess communication was used between modules running on the host computer, whilst USB-serial protocols were used to communicate between the software and firmware. The communication layers between the various components is outlined in figure 15; interested readers should consult the provided source-code for further details.

### S.2.1  Robot Controller

The robot controller functions to translate experimental procedures form a high-level description into a G-code representation. The G-code can be considered as an intermediate layer between the description of a recipe and the digital actuation, effected by the electronics. By using this translation pipeline, a number of expert-written modules could be leveraged, significantly reducing development costs and time. The core component of the RC is the PrintRun[4] library, developed for controlling RepRap printers (via G-code) and and which constitutes the most popular choice of library for this purpose. As this central library is written in the Python programming language[5], this language was used for the entire development of this layer. PrintRun communicates with the Arduino-based Firmware via a USB emulated-serial connection, facilitated by the pySerial library on the host computer[6].

The RC's interface to exterior code was encapsulated within a library called "RobotCtl". This library contains functions which relate to the fixed set of modular lab operations (e.g. "form droplet" or "clean petri dish") required to perform all experiments. The interface functioned to convert a sequence of these operations into a sequence of G-code instructions. The robot could move the syringes to any position around X or Y with a precision of 0.1 mm and this precision was accounted for in the conversion. Since there were multiple components on the X-Y carriage, the relative distance between these components was hard coded and used to modify the final X-Y position when one specific apparatus was selected. In contrast to the operational mode of a 3d-printer, which produces the entire batch of G-code from a 3d model in one run, the RC runs in online mode, compiling operations into G-code as they are received. This mode of operation is a necessity, due to the iterative nature of the procedures, whereby future operations are unknown until data from present operations have been processed. The Gcode produced in real-time is communicated to the robot via the PrintRun core.

### S.2.2 Firmware

The firmware layer is directly responsible for the actuation of mechanical parts. The firmware was written specifically for the target Arduino boards, hence the Arduino development environment was used for this layer. The native language of this environment is C++; therefore, this was the language used to develop the firmware layer.

| Code | Description |
| --- | --- |
| PX | Selects pump X. $X \epsilon\ [0...6]$ |
| MY | Selects motor Y. $Y \epsilon\ [0, 1]$ |
| DZ | Selects plunger direction. $Z \epsilon\ [0, 1]$ |
| SA | Sets speed. A is the number of ms of wait between steps |
| EB | Sets number of steps. Pumps used are limited to a maximum of 50000 steps |

Table 1: "PX MY DZ SA EB" Example of G-code operations output by the robot controller.

There were two separate Arduino boards, one to control the X-Y carriage (subsection S.1.1.5) and one to control the fluid platform (subsection S.1.1.6) As with the RC layer, the carriage firmware was built on top of extant 3d-printer modules. In this case, the firmware core used was the Sprinter package[7], commonly used with Reprap 3D printers. Functionality with regards to Z-axis movement, temperature control and extrusion of thermoplastic was removed from the code-base. In its stead, functionality was added for control of the syringes (See subsection S.1.2.3), via the servo library provided by Arduino[8]. The modified Sprinter firmware was therefore capability of X-Y motion, able to actuate all servo-motors and synchronize these actuations with the RC component, via the receipt and parsing of G-code instructions.

The fluid platform firmware was developed *de novo*. Each pump contained two components: A three-way valve and lead-screw actuated plunger. Both of these components sourced motive force from a NEMA stepper motor. These motors were controlled, as with other motors on the robot, via an "A4988 Stepper Motor Driver". A G-code interface was added, with new symbols for pump movement (Table 1). Very little sanity checking was performed and so the RC was relied upon for coherent operation. Figure 16 outlines the components described.

## S.2.3 Computer Vision

The computer vision component functioned to evaluate the behaviours, exhibited by the droplets, for further analysis by the AI component. The robot used a "PS3 EyeToy" as the basic sensor, to record video data from the under-side of the experimental arena. To facilitate visual analysis, a white background panel was used to cover the petri dish, to block ambient visual noise. A resolution of 640x480 pixels and a frame-rate of 30 FPS was used for video recording. The core component of the computer vision component was the library OpenCV (version 2.4.7)[11], with SciPy[12] used for some additional analysis. All analysis was performed using the Python bindings for OpenCV.

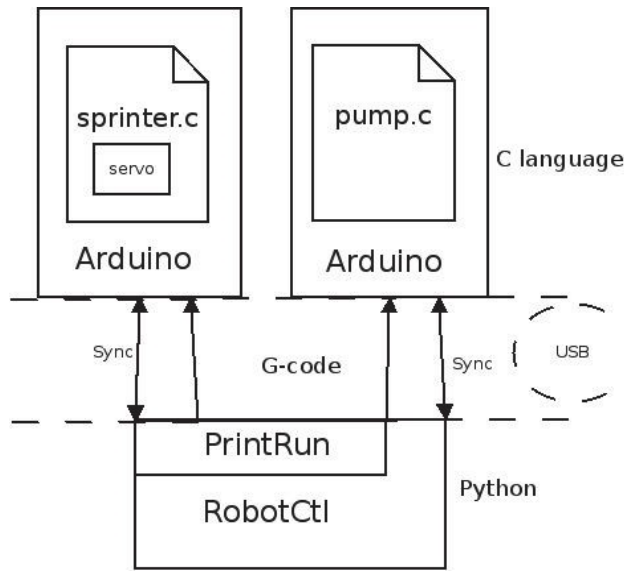

Figure 16: Software architecture describing the main components. The robot used two Arduino boards running a different firmware. One extends the 3D printing functionality with liquid handling capabilities. The other controls a set of pumps. The communication with the host-based python RC was performed using a serial protocol through USB.

### S.2.3.1 Common Image Processing

The CV component consisted of multiple image processing pipelines, one for each behaviour that was to be evaluated. Through this pipeline, droplets were identified in raw frame data (Figure 18, top left) and analysed for their position, size, shape and colour. Raw frame data was presented to the image processing pipelines in the form of RGB images, as received from the camera. No prior information was used to inform the image analysis pipelines. Figure 17 outlines the complete processing pipeline.

The initial step for all pipelines began with a Hough transform [13], used to detect the petri dish. Once detected, an analytic arena, slightly smaller than the petri dish, was defined. The aim of this reduction was to remove those droplets that had become caught against the edge of the dish; these droplets were considered "dead".

After the Hough transform, the pipeline was split into two parallel tracks. One of these tracks, aimed at long-term analysis, was based on foreground subtraction 18. The objective of this track was the discovery, and removal, of ambient background from the chemical components of the experiment (droplets). Because a white background was used, only the arena, with aqueous phase present, and droplets were present. The second track was targeted at analysis of early frames and was based on a chain of image processing operations. The objective of the image processing track was the discovery of closed structures, which corresponded to droplets. These structures could have any shape, as long as they could be represented by a connected component.

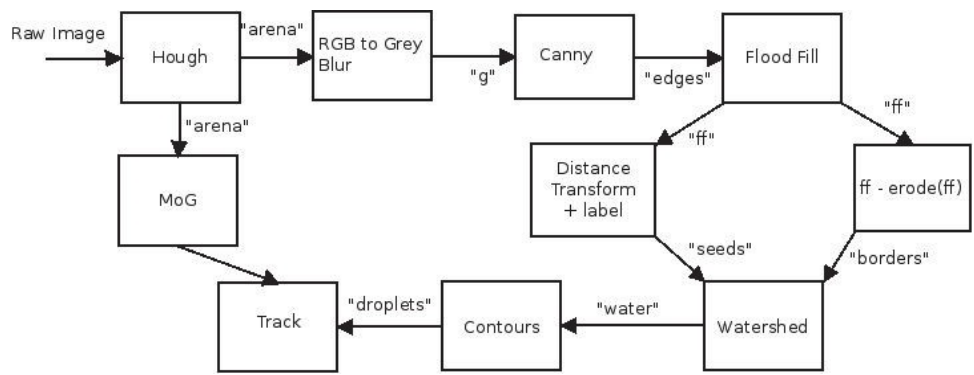

Figure 17: Scheme outlining the pipeline of the different techniques and algorithms involved in the droplet detection.

The image processing track began with an RGB-to-grey blurring operation, applied to remove noise. This was followed by the application of the Canny edge detection algorithm[14], resulting in an edge map (Figure 18, top right). As contours may be missing pixels (as can be seen in figure 18), a dilation operation was applied to fill these gaps. This was followed by a "flow fill" operation, with the origin at pixel (0,0) (Figure 18, bottom right). The result of this operation was the removal of any non-closed structure. The previously calculated Hough transform was then applied, to define the operational arena (Figure 18, bottom right). The next step in the pipeline was the application of a distance transform, to remove noise and to disconnect droplets that may have been artificially connected into one structure as a result of the dilation operation (Figure 19, top right). These final connected components were labelled using SciPy; the labels and the borders of the connected components were sent to a watershed algorithm[15], to recover the original size of the droplets (Figure 19, bottom left). This concluded the image-processing track of the parallel CV.

The foreground reduction algorithm used a mixture of Gaussians[16]. The default OpenCV parameters were used for fitting the mixture model, with learning rate set to 0.05. The result of the mixture model was a foreground (droplets) with background information removed.

The two parallel tracks resulted each in a binary image, where pixels were either 'on' (pixel is part of a droplet) or 'off' (no droplet). These two images were combined, using a boolean "or" operation to give definitive droplet-only images. The contours were then discovered on this combined image (Figure 19, bottom right). With this contour map, the area, size, shape, colour and centre of the droplets could be calculated. These data were then passed on to the fitness-specific components of the image-processing pipeline.

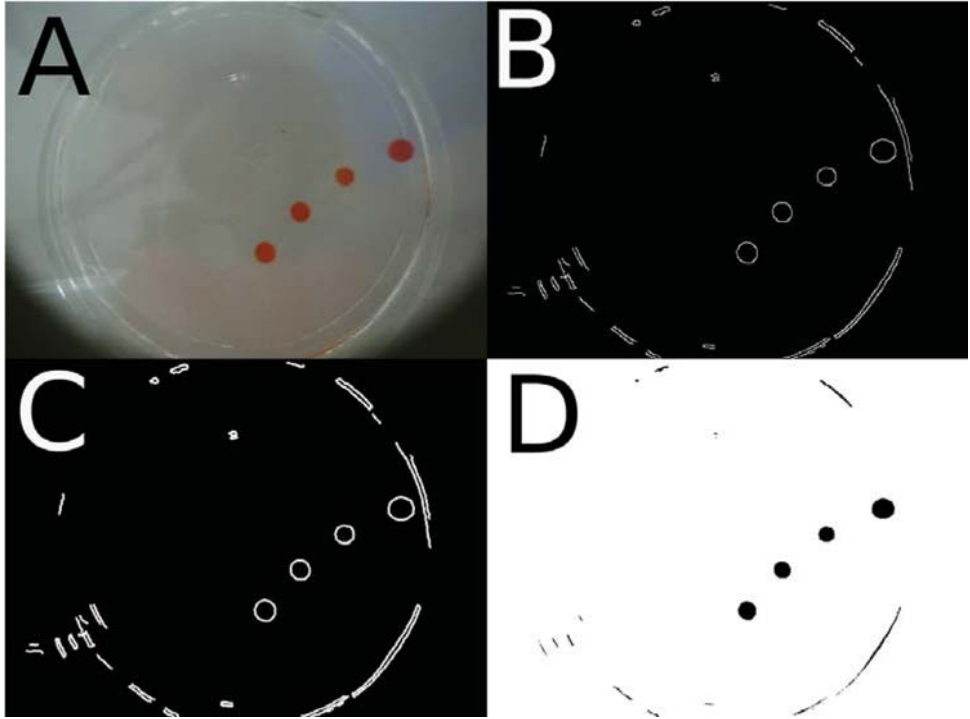

Figure 18: Droplet detection, image processing pipeline. Top left: Raw frame. Top Right: Canny edge detection. Bottom Left: Dilate morpho operation. Bottom Right: Flood fill operation, seed at pixel 0,0.

### S.2.3.2 Tracking

A common way to track objects is to use a filter, like the Kalman filter[9] or a particle filter[10]. Since the camera produced 30 FPS, overlapping consecutive frames to track droplets over time was enough considering the size of a droplet and how much it could move between frames.

Given a droplet $d_t$ in frame $t$, and a set of droplets $D_{t-1}$ in frame $t-1$, a set of candidates $C_{t-1}$ was built as the droplets in $D_{t-1}$ whose center was inside an area defined as a circle with 30 pixels of radius around $d_t$. The best candidate in $C_{t-1}$ was chosen as it was the nearest droplet to $d_t$ using the Euclidean distance. If there were no droplets inside this area, or all the droplets were already assigned to another droplet, this droplet was considered new.

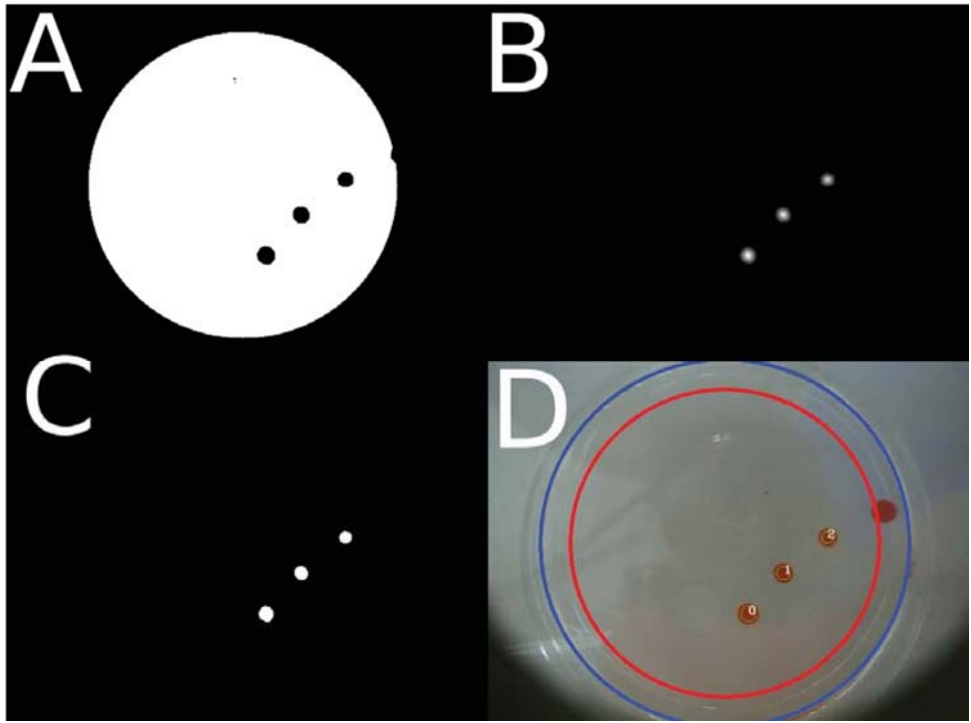

Figure 19: Droplet detection, image processing pipeline. Top left: Hough transform, dish detection. Top Right: Distance transform. Bottom Left: Watershed algorithm. Bottom Right: Final result. Blue circle marks the dish detection, red circle represents the arena.

### S.2.3.3 Experimental Data Generation

For each experiment, a data structure describing the positions of the droplets over time was produced. This data was used in order to rate an experiment based on different factors. During this research, the behaviours analysed in this way were "division", "movement" and "directionality".

### S.2.3.4 Division

Every experiment began with the robot placing droplets in the aqueous medium. Depending on the chemistry, the droplets could split as soon as they interfaced with the aqueous phase, but the robot made exactly four injections.

Division was defined as the number of droplets alive at the end of the experiment. We considered as viable any droplet with an area greater than 15 pixels. The experiment aimed to find droplet recipes that would divide in a controlled fashion, producing viable offspring rather than disparate fragments.

### S.2.3.5 Movement

Given a droplet $d$, its movement was defined as the Euclidean distance described by its translation between frames $t$ and $t + 1$. This translation was described in pixels as the fundamental units of distance.

Since between a pair of frames there can be several droplets moving, the total sum of distances described by all the droplets was divided between the number of droplets, obtaining the average distance translated per droplet. Every experiment ran for 1 minute, containing a few thousand of frames. The average distance per droplet was calculated for every pair of frames, its quantity summed, and then divided by the total number of frames, in order to obtain the average translation described per droplet per frame.

The movement-derived fitness is then given by

$$W_{movement} = \frac{1}{MN} \sum_{t=0}^{N} \sum_{i=0}^{M} \sqrt{(x_{i,t} - x_{i,t-1})^2 + (y_{i,t} - y_{i,t-1})^2} \quad (1)$$

where *N* is the total number of frames in the video sequence, *M* is the total number of droplets observed and where $(x_{i,t}, y_{i,t})$ are the Euclidean coordinates of droplet *i* on frame *t*

### S.2.3.6 Directionality

Given a droplet *d* in frame *t*, *t* + 1 and *t* + 2, its position for each frame in the *XY* plane is denoted by the points *A*, *B* and *C*. Two vectors were defined: $\vec{v}$ as $\vec{AB}$ and $\vec{w}$ as $\vec{BC}$.

By directionality of a droplet it is meant the angle between *v* and *w* which represents the change of direction on a droplet moving pattern. Low values map to droplets which move in straight lines, middle values to droplets which describe curves and high values droplets that vibrate or wobble. Inverting the dot product formula:

$$\vec{v} \cdot \vec{w} = \|\vec{v}\| \|\vec{w}\| \cos \alpha \quad (2)$$

The angle is obtained:

$$\alpha = \arccos\left(\frac{\vec{v} \cdot \vec{w}}{\|\vec{v}\| \|\vec{w}\|}\right) \quad (3)$$

For each experiment, the angular rotation, per droplet, per frame was given was used as the directional fitness, as given by

$$W_{directionality} = \frac{1}{MN} \sum_{t=0}^{N} \sum_{i=0}^{M} \alpha_i \quad (4)$$

### S.2.4 Artificial Intelligence

For each behaviour to be tested, three Genetic Algorithm (GA) runs were used. Each GA run performed 21 generations, with a fixed-population size of 25 individuals and 15 individuals being propagated from the previous generation, for a total of 225 recipes. Each recipe was repeated three times, and the minimum between the mean and the average of these 3 experiments was returned. Each experiment generated four droplets.

A complete test ran the GA 3 times, therefore in total it generated 675 recipes. Each recipe was repeated 3 times, for a total of 2025 experiments. Each experiment generated 4 droplets, for a total of 8100 droplets.

Individuals were fixed-length genomes of 4 floating-point numbers (i.e. Quantitative trait loci). The GA used a per-locus probability of mutation (resulting in a poisson-distributed number of mutations per individual). For each locus selected for mutation, a normal-distributed noise function, with a mean of 0 and an SD of 0.1 was additively applied. Each child was always the product of a single crossover recombination between two distinct parents, with the crossover being uniformly distributed along the genome and the same genetic location being used for each parent. Individuals were birthed with parents being selected, without replacement, from the extant pool with probability proportional to the fitness (to the power of some parameter). After birthing and fitness measurement, the population was culled to a fixed size, with individuals being chosen for death with probability inversely proportional to the fitness (to the power of some selective pressure parameter).

| Parameter | Value |
| --- | --- |
| Generations | 21 |
| Genome length | 4 |
| Population size | 25 |
| Carry-overs | 15 |
| Per-locus mutation rate | 0.3 |
| QTL mutation (SD) | 0.1 |
| Selective pressure | 1 |

Table 2: Parameters used to generate all GA-derived data presented in this paper.

## S.3 Chemistry

Oils and surfactants were purchased from Sigma-Aldrich and used as received, unless otherwise stated. All experiments were performed at 22 degrees Celsius.

### S.3.1 Preparation of aqueous phase

Tetradecyltrimethylammonium bromide (TTAB) (6.73 g, 20.0 mmol) was dissolved in distilled water (*ca* 600 mL), adjusted to pH 13.00 with 5 M NaOH solution and made up to 1 L to give a 20 mM solution at pH 13.00.

### S.3.2 Preparation of oils

The oils (1-octanol, octanoic acid, dodecane, 1-pentanol and diethyl phthalate) were prepared in 200 mL aliquots in reagent bottles. Each oil was dyed with 0.25 mg/mL Sudan III and vortexed to mix.

### S.3.3    Cleaning cycle

After each experiment, acetone (*ca* 3 mL) was pumped into the petri dish to dissolve remaining oil droplets and the mixture was aspirated from the dish to the waste container. The dish was then washed with acetone (2 x 3 mL) and aqueous phase (2 x 3 mL) Less thorough cleaning cycles were often to leave trace residues, which interfered with subsequent experiments.

## S.4    Methods

A fully automated liquid handling robot capable of producing droplets in a Petri dish with an aqueous sub-phase was constructed. The robot was based upon a RepRap 3D printer architecture with a webcam for video recording / image analysis.

| Oil | Density ($gmL^{-1}$, 20 °C) | Value Ref. | Solubility ($gL^{-1}$) | Value Ref. | Surface Tension ($mNM^{-1}$) | Value Ref. | Viscosity ($mPas^{-1}$) | Value Ref. |
|---|---|---|---|---|---|---|---|---|
| 1-octanol | 0.824 | [23] | 0.46 | [17] | 27.1 | [17] | 7.288 | [17] |
| 1-pentanol | 0.811 | [23] | 22 | [17] | 25.36 | [17] | 3.619 | [17] |
| DEP | 1.12 | [23] | 1.08 | [17] | 19.6 | [19] | 10.625 | [22] |
| Octanoic Acid | 0.91 | [23] | 0.68 | [18] | 27.9 | [20] | 5.020 | [17] |
| Dodecane | 0.78 | [23] | Insol. | [18] | 25.35 | [21] | 1.383 | [17] |

Table 3: Physical properties of oils used in the experiment

The droplet formulations are produced in well-plates (96-well format) where each well is stirred with a magnetic stirrer bar. The droplet formulations are based upon the following reagents ((1-octanol, 1-pentanol, diethyl phthalate (DEP), dodecane and dilutions of octanoic acid in one of the other oils (typically 20The formulations were delivered from the computer control in the form of four numbers, which represented the proportion of each oil; therefore the total summed to "1.0".

The total volume of the oils place in a well was 360 $\mu L$. The specific quantity for each oil was this volume multiplied by their relative proportion.

Once the mixing was completed, three experiments were conducted from the final formulation. For each experiment, 80 $\mu L$ of oil were absorbed using the syringe. Of this volume, four 5 $\mu L$ oil droplets are then placed in the aqueous phase (20 mM aqueous tetradecyltrimethylammonium bromide (TTAB), adjusted to pH 13.00 with 5 M NaOH solution). These positions were consistent between experiments, with pre-programmed X-Y coordinates. The needle was laid just above the aqueous phase and the droplets were released to the surface of the liquid. Once a droplet was outside the needle, the syringe was moved up to its default position. This movement broke the tension between the droplet and the needle, depositing the droplet over the aqueous phase. This process is summarised in Figure 20.

A video of the resulting droplet behaviour was then recorded from beneath the dish using a camera at a resolution of 640 x 480 pixels at 60 fps from below after covering the top of the Petri dish with a white background for 60 seconds. After the experiment the entire contents of the dish are automatically removed, cleaned with three washes of acetone following with three washes of aqueous phase. During the acetone wash, the syringe is cleaned also with acetone. After a complete generation has concluded, the evolutionary algorithm quantifies the fitness of the formulations and calculates the next set of reagent volumes. Image analysis of the droplet behaviour against the fitness function provides the next set of inputs for the evolutionary algorithm.

Our four component system comprises 1-Octanol, Diethyl-phthalate, 1Pentanol, and either Octanoic acid or Dodecane as the fourth ingredient. Movement of 1-pentanol droplets is described in the literature [30], and a range of other mid-to-long chain alcohols were tested using automation to explore a range of solubilities. Diethyl phthalate is known to form stable droplets[31]. Dodecane was chosen as an extremely non-polar oil with low density and viscosity. Octanoic acid was explored as it is deprotonated at high pH to form an anionic surfactant.

Once the experiment was concluded, the syringe pumps were used to pump acetone and aqueous phase into the Petri dish and then to remove the solvated waste. First, three washes of acetone were performed: 4 ml, 4 ml and 3 ml. During the first two of these washes, the needle was dropped into the acetone, and the plunger was actuated to absorb and release acetone, cleaning the needle and syringe.

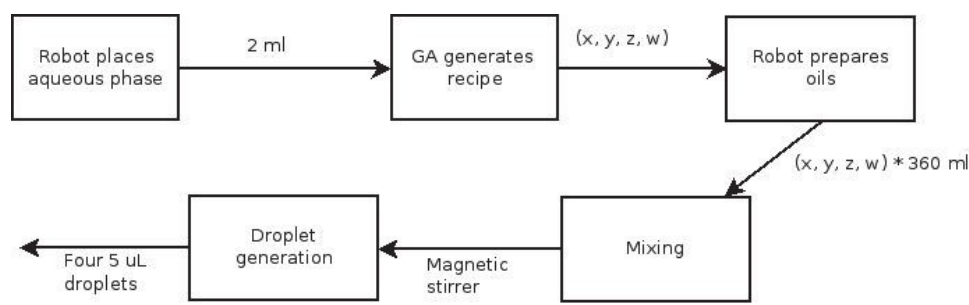

Figure 20: Outline of the droplet generation. First the aqueous phase is placed, and then the droplets are dropped over it.

After washing with acetone, three washes with aqueous phase were performed, with 1.5 ml, 1.5 ml and 1 ml. The objective of the aqueous phase wash was to remove the remains of acetone from the petri dish so that it would not interfere with subsequent experiments. This process is summarized in figure 21.

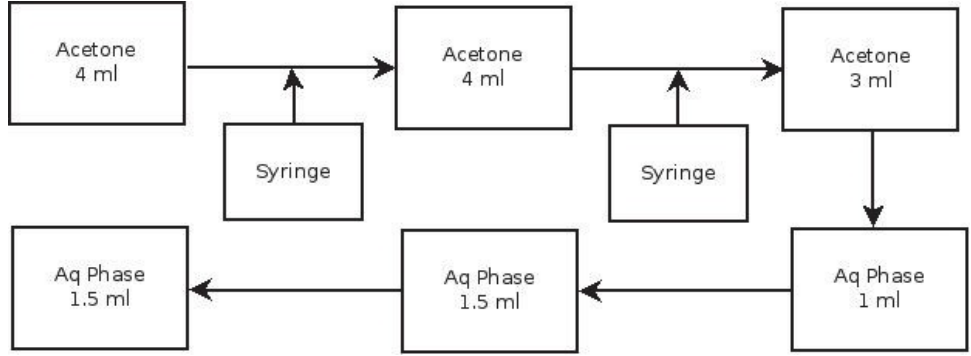

Figure 21: At first, three washes with acetone were performed. Between the first and second, and the second and third, the syringe is dipped into the acetone, and the plunger actuated, cleaning the needle, barrel and plunger. The three following aqueous phase washes were performed to remove the remains of acetone.

## S.5 Analysis

### S.5.1 Processing of Fitness Landscapes

The fitness landscapes seen in Figure 6 in the main text were produced through a multi-step analytical pipeline. The results from the GA were first collated and passed through a radial basis function kernel ridge regression to produce a model. This model was then queried through a grid search along each face of the parameter-space simplex to estimate the fitness at each location.

#### S.5.1.1 Kernel Ridge Regression

General Linear Regression (GLR) performs model fitting by minimizing the sum-of-squares error over a space of linear coefficients, in an equation of the form

$$\widehat{y}_i = \widehat{\boldsymbol{\phi}}^\top \boldsymbol{x}_i, \tag{5}$$

with the sum-of-squares error being given for $N$ points as.

$$(\widehat{\boldsymbol{y}} - \boldsymbol{y})^\top (\widehat{\boldsymbol{y}} - \boldsymbol{y}) \tag{6}$$

Ridge-regression, also known as Tikhonov regularisation[24], is a commonly used method of regularization of ill-posed problems. This form of regression augments the minimization with a weighted penalty on the size of the coefficient vector φ:

$$\widehat{\phi} = \arg\min_{\widehat{\phi}} (\widehat{y} - y)^\top (\widehat{y} - y) + \lambda \widehat{\phi}^\top \widehat{\phi} \tag{7}$$

Kernel ridge-regression is a re-formulation of ridge-regression, used in situation where the number of dimensions exceeds the number of data-points. Equation 5 is substituted by

$$\widehat{y}_i = \boldsymbol{\theta}^\top \boldsymbol{X} \boldsymbol{x}_i \tag{8}$$

With equation 7 taking the equivalent substitution:

$$\widehat{\phi} = \arg\min_{\widehat{\phi}} (\widehat{y} - y)^\top (\widehat{y} - y) + \lambda \widehat{\boldsymbol{\theta}}^\top \widehat{\boldsymbol{\theta}} \tag{9}$$

An important property of this reformulation is that it requires the calculation of a dot-product between each input vector and every other. These dot-products can be collated in the Gram matrix:

$$\mathbf{G} = \mathbf{X}\mathbf{X}^\top \tag{10}$$

However, the dot-product does not need to be computed in input space. Higher predictivity can often be obtained by transforming the input vectors into a higher-dimensional space, known as "feature space":

$$z_i = f(x_i) \tag{11}$$

the Gram matrix G is then replaced by the kernel matrix K, whose entries are given by

$$K_{ij} = \boldsymbol{z}_i^\top \boldsymbol{z}_j. \tag{12}$$

Rather than calculating the dot products through an explicit mapping from input space into feature space, it is usually quicker to use an implicit feature-space inner product $g$, calculated directly on the input vectors:

$$K_{ij} = g(x_i, x_j) \tag{13}$$

The function $g$ is then known as the "kernel function" and its use, with associated improvements in computational speed, is often referred to as the "kernel trick".

### S.5.1.2  Radial Basis Function

The kernel function that we apply to extend the sampled points into a smooth fitness landscape is the Gaussian Kernel. This kernel function falls into the broader category of Radial Basis Functions (RBFs), whose members are defined by being real-valued function that depend only on the distance, in input space, between the input vectors. Distance is here defined between points $i$ and $j$ as

$$r_{ij} = \|\mathbf{x}_i - \mathbf{x}_j\|. \tag{14}$$

The Gaussian RBF then defines the kernel function between two points $i$ and $j$ as

$$K_{ij} = e^{-\frac{r_{ij}}{2\sigma^2}}. \tag{15}$$

Where $\sigma$ is a problem-specific parameter, roughly equivalent to the standard deviation of the Gaussian probability distribution. One peculiar property of the Gaussian kernel function, not shared by other RBFs, is its implicit mapping of the input vectors into an infinite dimensioned feature space. This property arises from the expansion of the exponential term, here demonstrated for a single-dimension input vector:

$$K_{ij} = e^{-\frac{(x_i-x_j)^2}{2\sigma^2}} \tag{16}$$

$$= e^{-\frac{x_i^2}{2\sigma^2}} e^{-\frac{x_j^2}{2\sigma^2}} \sum_{k=0}^{\infty} \left[\frac{2^k x_i^k x_j^k}{2\sigma^2 k!}\right] \tag{17}$$

As a result, whereas other kernel functions can be used to derive a set of feature space coefficients, the Gaussian RBF requires that the kernel function be evaluated for every *N* training vector with a new vector, for a prediction to be made for that vector:

$$\widehat{y}_k = \sum_{i=0}^{N} K_{ik}\theta i \tag{18}$$

### S.5.1.3  Complete Landscapes

Three reduced fitness landscapes are shown in the main text in figure 6. Each of these subfigures shows a three-dimensional "facet" of the four-dimensional space of oil composition. Each facet was chosen so that the global maximum, for each environment, would be shown, in each case. However, as each facet is derived by holding one of the oil proportions at zero, there are three other facets per environment. All facets for the three environments are therefore shown below.

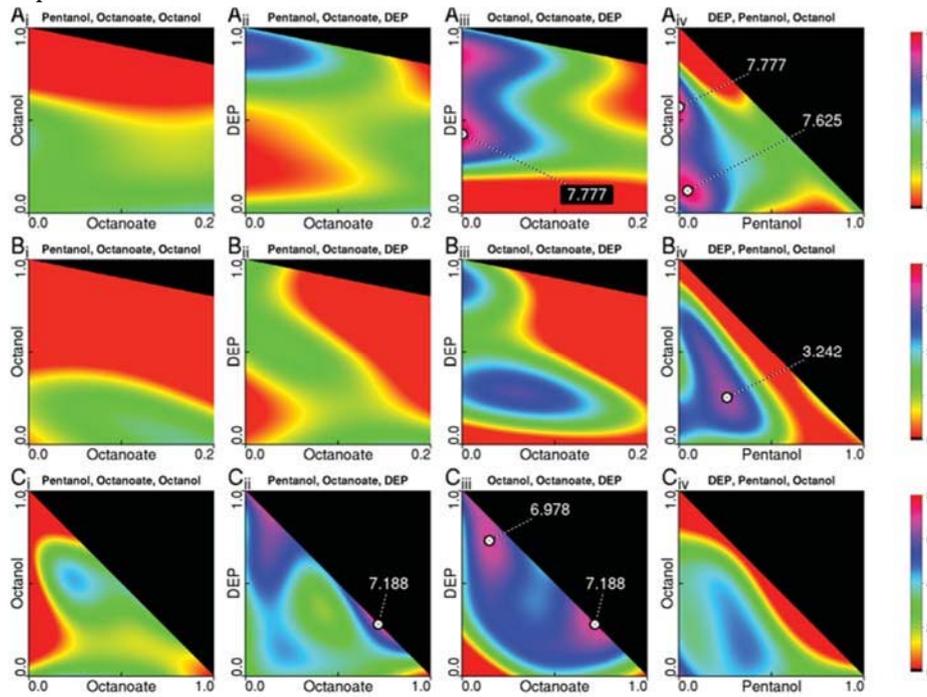

Figure 22: Fitness landscapes for the environments division (**Ai-iv**), motility (**Bi-iv**) and vibration (**Ci-iv**. In each plot, three substances are displayed as the plot title. The fourth substance can therefore be assumed to be held at a constant zero. The projection used is equivalent to the ternary plots shown in the main text. Each axis shows the proportion of a substance (indicated). The proportion of the third substance (**Z**) is calculated as $Z = 1 - X - Y$. The numerical indications show the location of fitness peaks. In division and vibration two major fitness peaks were discovered.

Minor fitness peaks are not indicated. The scale bar corresponds to the fitness functions for each environment and are not comparable between environments.

Figure 22 shows four fitness landscapes per environment, displaying a greater proportion of the analysis than that found in the main-text. As "true" representation of the data consists of a solid four-dimensional simplex, graphical representation is inherently difficult. The authors therefore opted to display on the faces of this simplex. This was considered a reasonable approximation as, for each environment, the global maximum was found on one of these faces and not internally. Indeed, for division and vibration environments, the larger of the two major fitness peaks was found on an edge between two faces (a two-substance composition). These maxima can therefore be seen, on two faces, for division in **Aiii** and **Aiv** (7.777) and for vibration in **Cii** and **Ciii** (7.188). Numerous sub-optimal local maxima are also discovered by the analysis in all environments and are explored in subsection S.5.1.4.

### S.5.1.4 Catchment

In an evolutionary fitness landscape, multi-modality corresponds to the concept of *fitness islands* [27]. Such a feature is defined as being that volume surrounding a local (or the global) maximum, such that for any point within that volume, consistent, upward progress along the gradient will converge at that specific maximum.

To analyse the number and boundaries of the fitness landscapes that underly the experimental evolution component of this manuscript, a discovery algorithm was run on the fitness hyperplanes derived through the kernel modelling. The hyperplanes were represented as 4 301 × 301 quantized lattices, as plotted in figure 4 in the main text. For each unique maximum, an active set was maintained, starting with a single location at that maximum. For each location in the active-set, all eight surrounding, quantized locations were tested, such that for each location tested, the 8 locations around *that* location were tested, to discover which of them corresponded to the neighbouring maximum. For each location whose neighbouring maximum was the current location from the active-set, that location was marked as belonging to the current fitness island and then added to the active-set for analysis of its own neighbourhood. It was simple to allow this search to extend over the boundaries of one hyperplane to another hyperplane, where the locations were equivalent (i.e. where one of the components was 0).

The results of this algorithm are presented in figure 23. For each of the three landscapes, exactly five fitness islands were observed, although there may be fitness islands, in the interior of the fitness space, not revealed by the search across the exterior hyperplanes. There is insufficient data to specify whether the consistency of the occurrence of five islands per landscape is a product of the dimensionality or a coincidental artefact.

### S.5.2 Evolutionary Trajectories

Statistical analyses were carried out on the evolutionary trajectory of each of the three optimization experiments. The raw fitness distributions, as a function of generation are shown in figure 24.

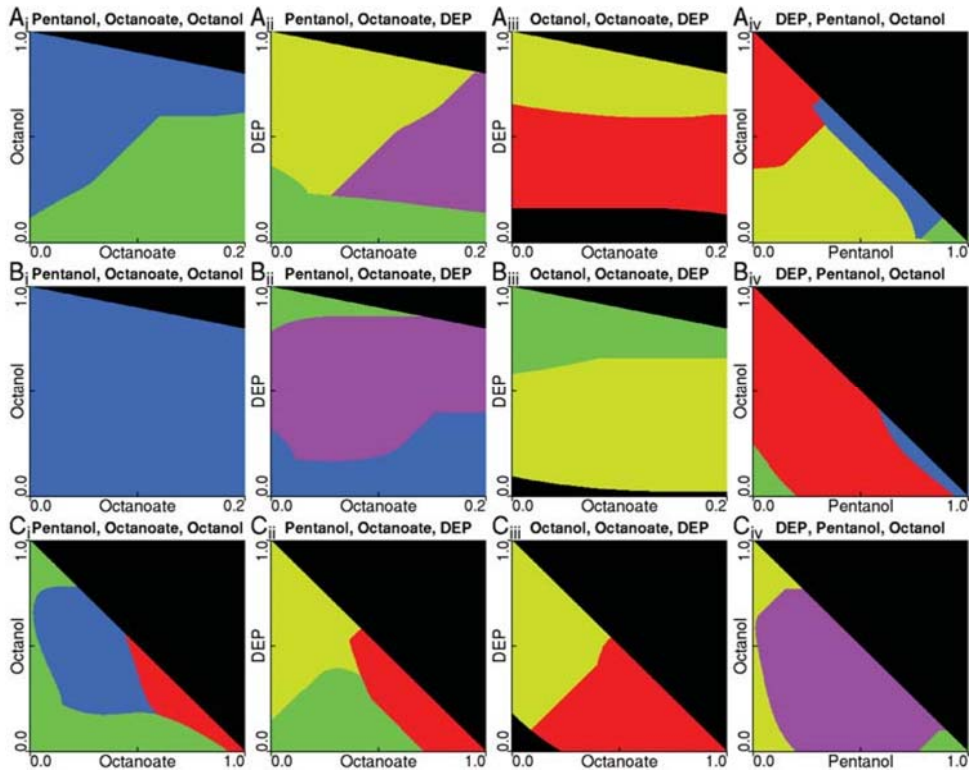

Figure 23: Fitness island map derived from the kernel analysis for the fitness landscapes division (**Ai-iv**), motility (**Bi-iv**) and vibration (**Ci-iv**). Each colour (Red, yellow, green, blue, purple) corresponds to a stable fitness island (colours stated in sorted order of island-maximum from highest to lowest [i.e. red is the fittest island and purple the least fit]). Five islands were observed in each landscape, although it is unknown if further islands are present in the interior of the full simplex.

It is stated in the introduction that the final population of droplets, in all cases, show enriched fitness compared to the initial populations. This analysis was performed by comparison of generation 1 with generation 21. It is common, in biological scenarios, to consider the fitter members of a population, rather than the average, since these will propagate. P-values for this analysis were therefore conducted by comparing only the top half (fitness greater than median) of each generation. This resulted in 38 individuals per generation (all three repeats were amalgamated). Analysis-of-Variance was then performed with fitness as a function of generation (21 or 1) and is summarized in table 4.

From the results presented in figure 24, it not visually obvious whether population fitness shows improvement during the latter half of each run. To determine whether this was the case or not, the same analysis as above was repeated, but with generation 21 compared against generation 11. As can be seen in the results presented in table 5, both division and vibration showed significant improvement from the halfway-point to the end. Motility, however, can be considered as having been optimized after the conclusion of 11 generations.

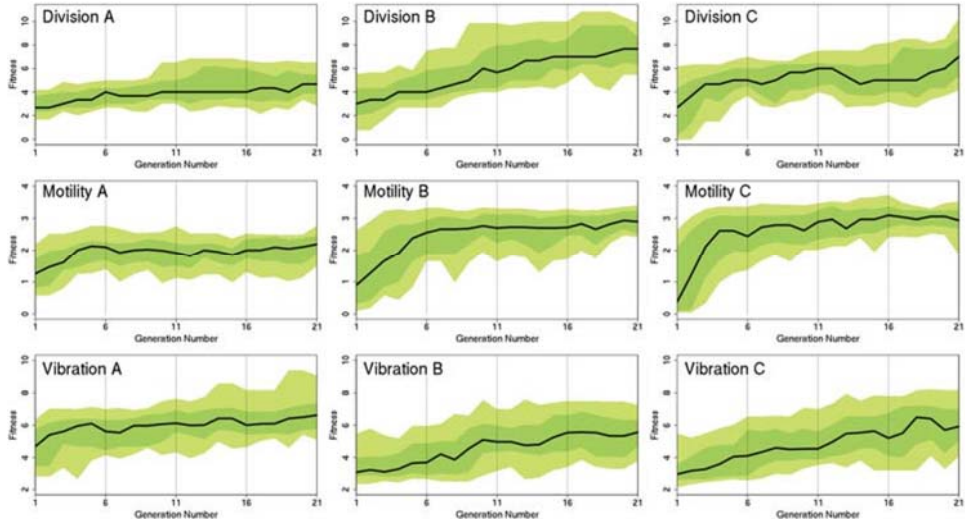

Figure 24: Fitness progression throughout each of the nine optimization experiments, derived from three repeats for each of the three fitness landscapes. Fitness is specific to each landscape and not comparable between landscapes. The black line corresponds to the median for each generation, dark green bounds the distribution between the upper and lower 25th percentile and light green bounds between the upper and lower 10th percentile.

It was then questioned whether there is a significant different in fitness variation as a function of generation. For this analysis, all generations were examined under an Analysis-of-Variance, comparing continuous fitness against categorical generation number. The results are summarized in table 6; all environments showed a highly significant result.

Finally it was tested whether there was a positive dependence of fitness upon generation (i.e. whether fitness increased positively with generation). Due to the non-linear association that was visually observed from figure 24, the Kendall correlation test was used[29]. The results are summarised in table 7; all environments showed a highly significant result.

### S.5.3 Lattice search

To better visually represent the behavioural space exhibited during the lattice-search, a self-organizing map analysis was applied to the resultant data. Readers unfamiliar with the technique are referred to [25] and [26] as introductory texts. Behaviours were manually assigned, based on visual assessment by one of the researchers. Each circle in figure 25 represents a "node": The fundamental unit of output from the SOM. Chemical composition varies across both the X and Y axes in a spatially significant, but visually non-obvious pattern. Each behaviour is assigned an individual colour and it can be seen that behaviour cluster together in space, and therefore in composition. Each cluster of behaviours therefore represents a phenotypic "island" within the composition/behaviour mapping.

| | Fitness | F-value | p-value |
|---|---|---|---|
| × | Division | 104.1 | $< 10^{-15}$ |
| | Motility | 74.9 | $1.7 \times 10^{-13}$ |
| | Vibration | 43.6 | $2.7 \times 10^{-13}$ |

Table 4: ANOVA analysis of the first generation individuals in the upper half of the fitness distribution against the last generation individuals in the upper half of the fitness distribution. All fitness landscapes showed highly significant improvement in fitness from the beginning, to the conclusion of the experiment.

| Fitness | F-value | p-value |
|---|---|---|
| Division | 5.51 | 0.0207 |
| Motility | 0.611 | 0.436 |
| Vibration | 7.08 | 0.00902 |

Table 5: ANOVA analysis of the $11^{th}$ generation against the last generation, under the same method of analysis as in table 4. Division and vibration both show significant improvement in fitness during the latter half of the experiment (Under the Holm–Bonferroni multiple testing correction[28]).

| Fitness | F-value | p-value |
|---|---|---|
| Division | 407.8 | $< 10^{-15}$ |
| Motility | 259.1 | $< 10^{-15}$ |
| Vibration | 297 | $< 10^{-15}$ |

Table 6: ANOVA analysis of all generations; the entire distribution for each generation was tested as a function of generation, expressed as a categorical variable. All fitness landscapes show highly significant differences in variation between generations.

| Fitness | $\tau$-value | Z-value | p-value |
|---|---|---|---|
| Division | 0.283 | 18.89 | $< 10^{-15}$ |
| Motility | 0.210 | 14.33 | $< 10^{-15}$ |
| Vibration | 0.256 | 17.50 | $< 10^{-15}$ |

Table 7: Kendall rank-correlation test of non-linear dependence of fitness on generation, expressed as a continuous variable. All fitness landscapes show a highly significant, positive correlation between the two variables.

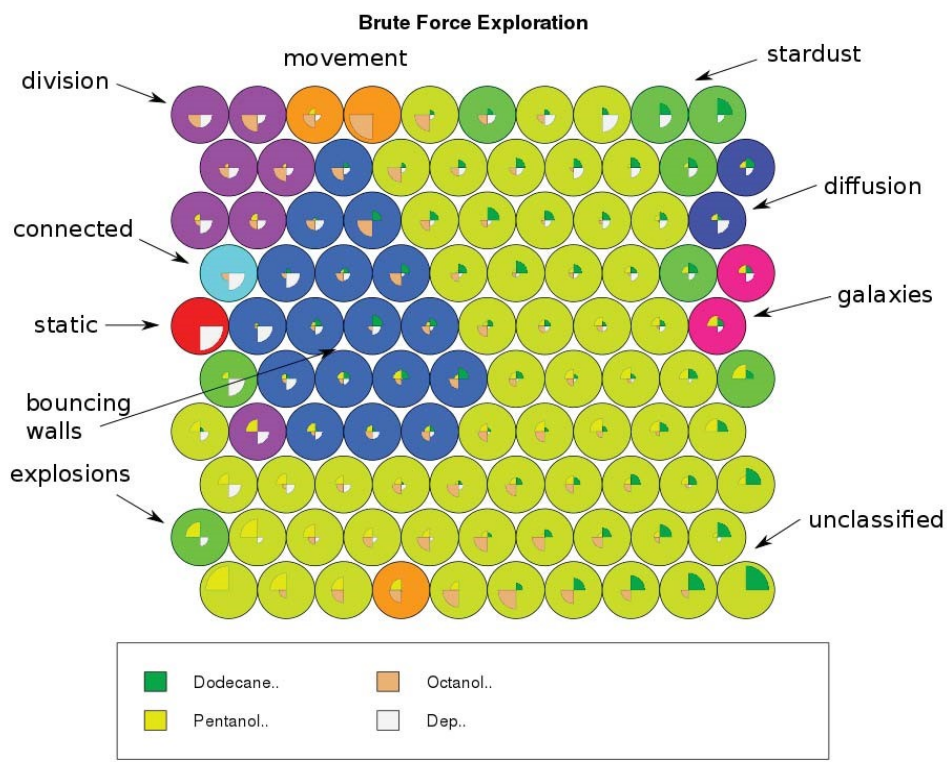

Figure 25: Self-Organizing Map of droplet behaviours across the formulation space. Each circle represents a resultant node, with colours assigned to individual behaviours. Chemical composition varies according to spatial coordinate with neighbouring coordinates having similar composition.

## S.6  Printed Parts

### S.6.1   Main frame

#### S.6.1.1    3DPP1, truss base

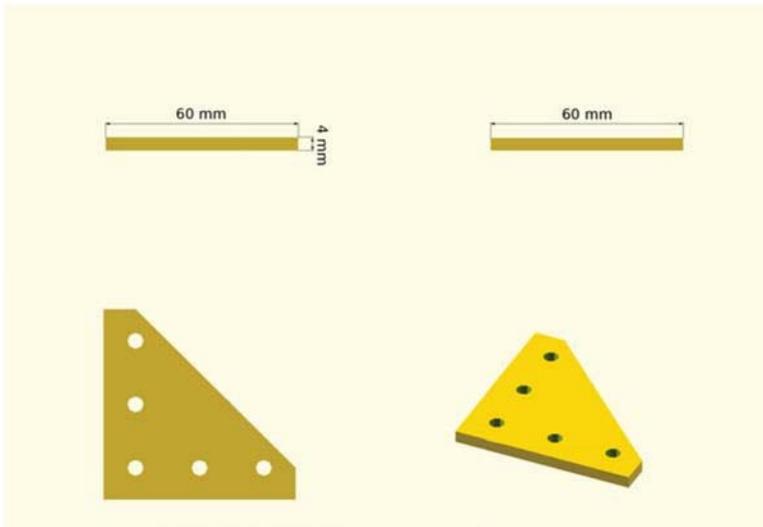

**3DPP1** Truss base for solid corner connection on the robot frame

#### S.6.1.2    3DPP2, truss foot

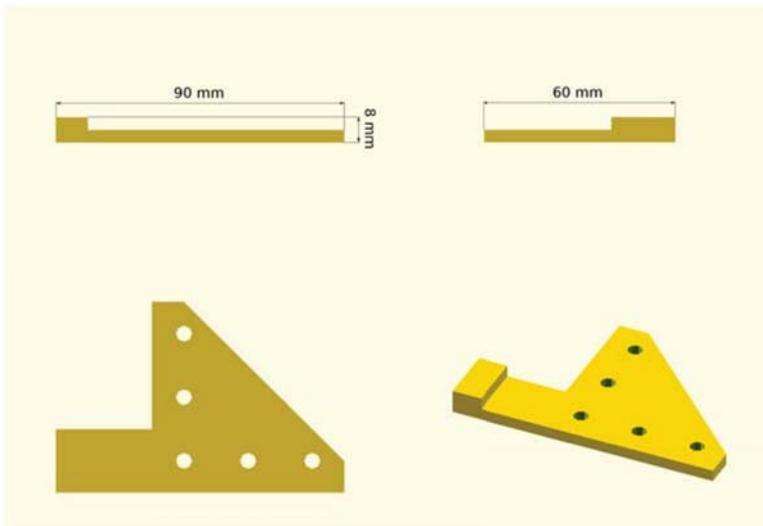

**3DPP2** Truss base and integrated foot for bottom half

## S.6.2 X-Y axis

### S.6.2.1 3DPP3, truss rod

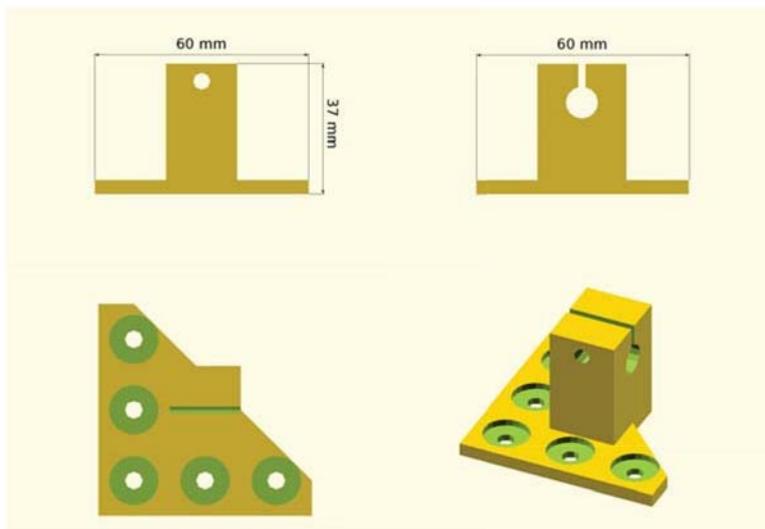

**3DPP3** Truss base and integrated rod holder for rounded steel bar

### S.6.2.2 3DPP4, truss motor

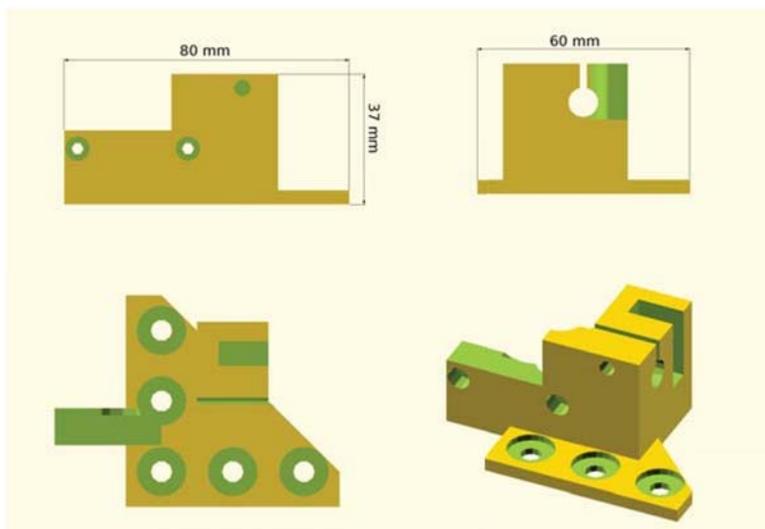

**3DPP4** Truss base and integrated motor holder for pulley drivers

### S.6.2.3   3DPP5, Y idler

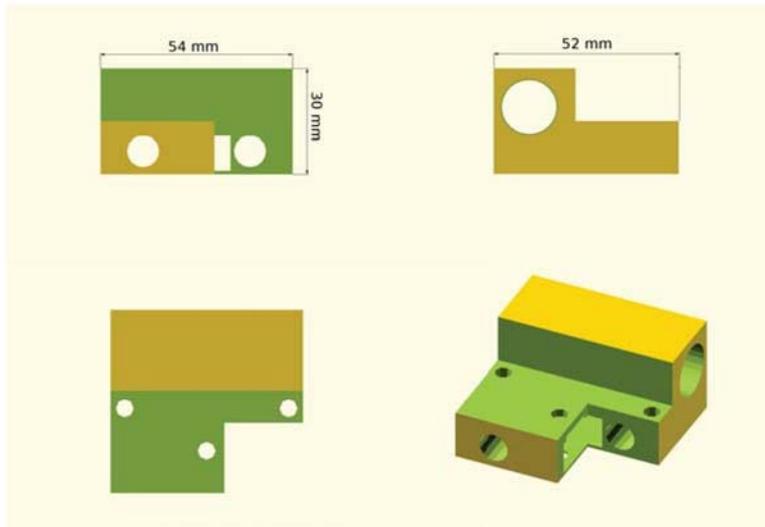

**3DPP5** X axis rail runner with attachment sockets for Y-axis rods

### S.6.2.4   3DPP6, Y idler with motor

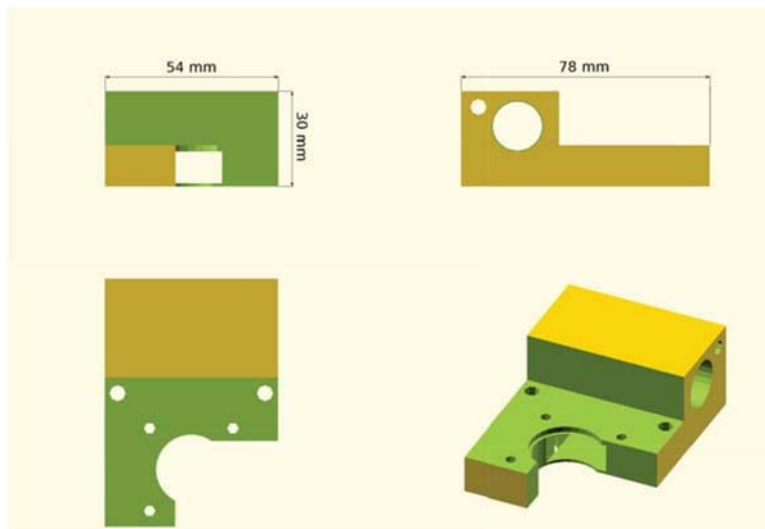

**3DPP6** X axis rail runner with attachment sockets for Y-axis rods and attachment point for Y motor

### S.6.3 X-Y carriage

#### S.6.3.1 3DPP7, carriage main component

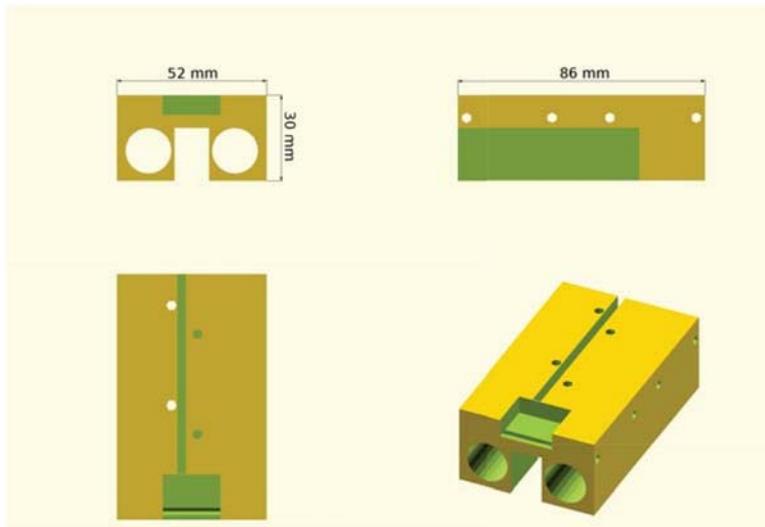

**3DPP7** Main carriage component, used as attachment point for other carriage apparatus

#### S.6.3.2 3DPP8, camera background panel

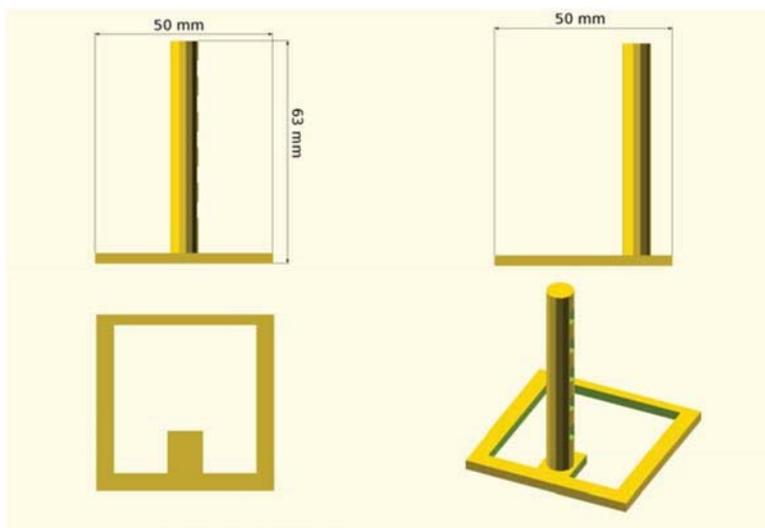

**3DPP8** Holder for white panel to provide uniform background the camera.

### S.6.3.3    3DPP9, panel support

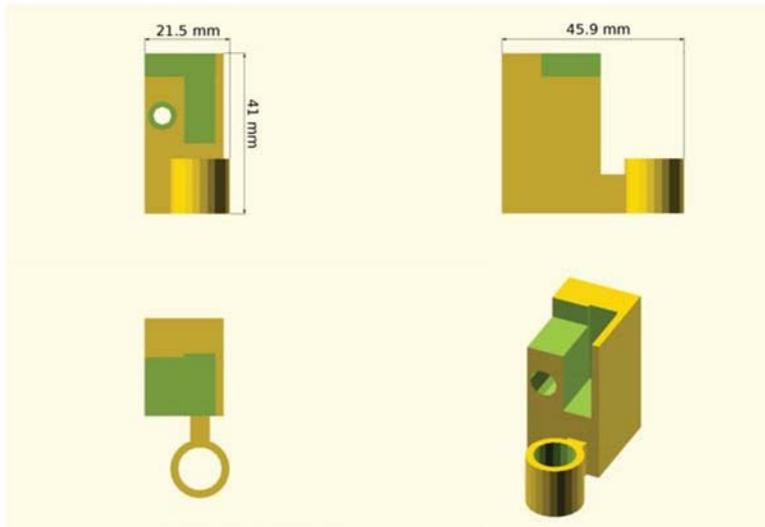

**3DPP9** Supporting component to steady the background panel.

### S.6.3.4    3DPP10, panel grip

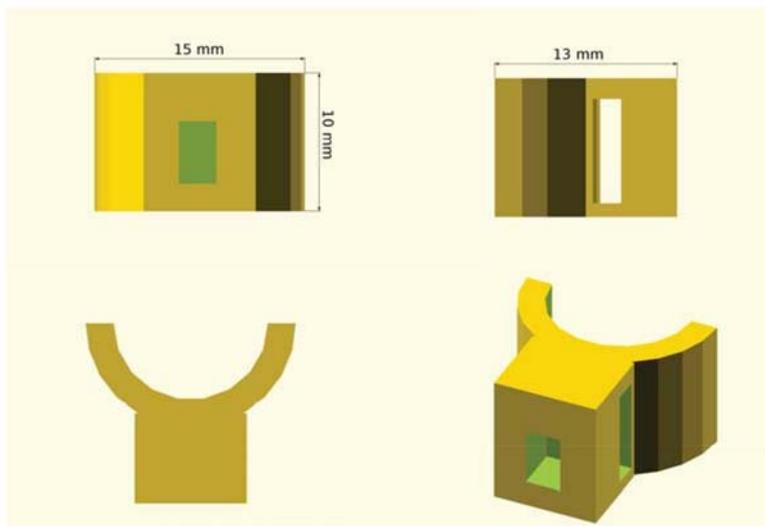

**3DPP10** Supporting component to steady the background panel.

## S.6.4 Fluid handling

### S.6.4.1 3DPP11, tube holder

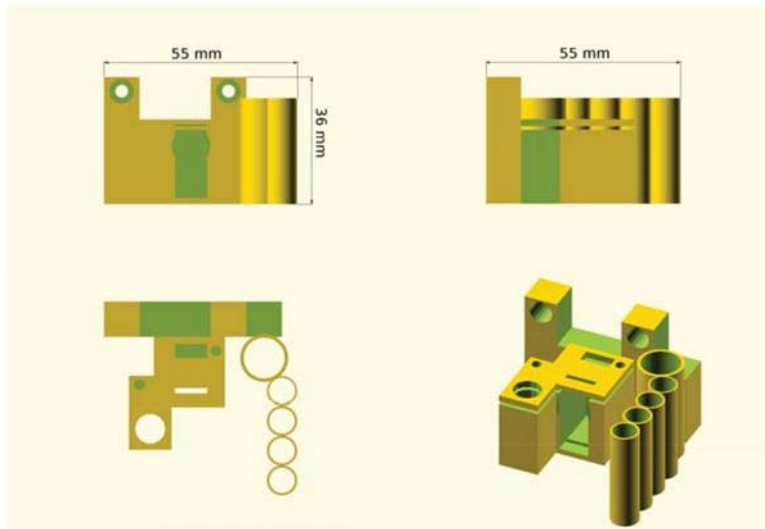

**3DPP11** Component for holding tubes incoming from the fluid platform, attaches to 3DPP7.

### S.6.4.2 3DPP12, sryinge holder

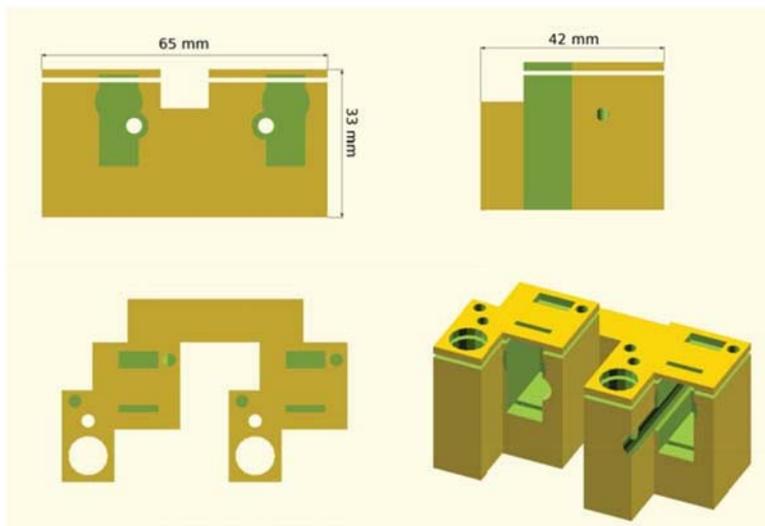

**3DPP12** Main component for holding syringes, attaches to 3DPP7.

### S.6.4.3  3DPP13, syringe holder cap

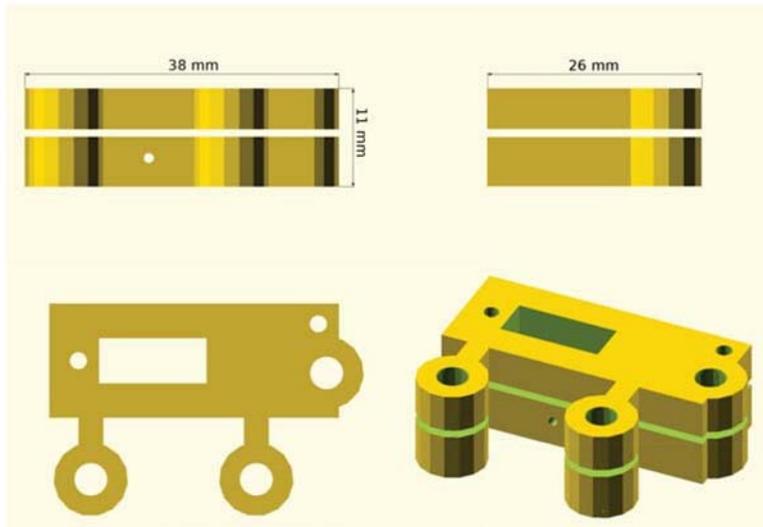

**3DPP13** auxiliary component for holding syringes, attaches to 3DPP11.

### S.6.4.4  3DPP14, plunger casing

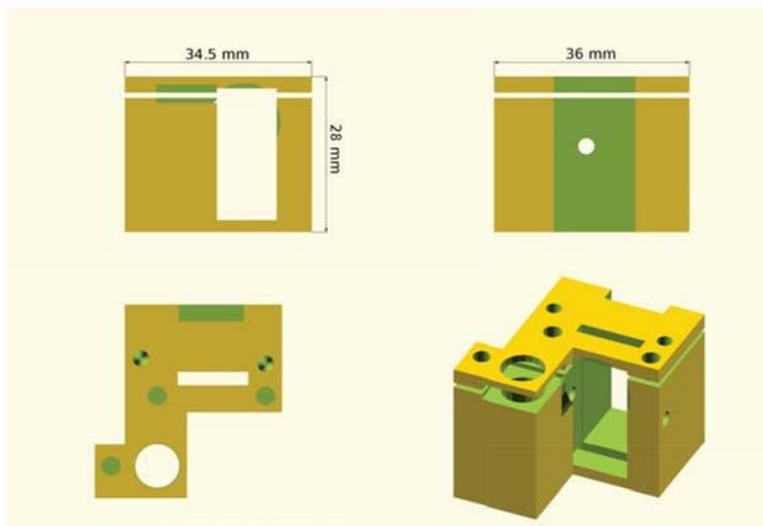

**3DPP14** component for holding and actuating the syringe plungers.

### S.6.4.5  3DPP15, syringe tip

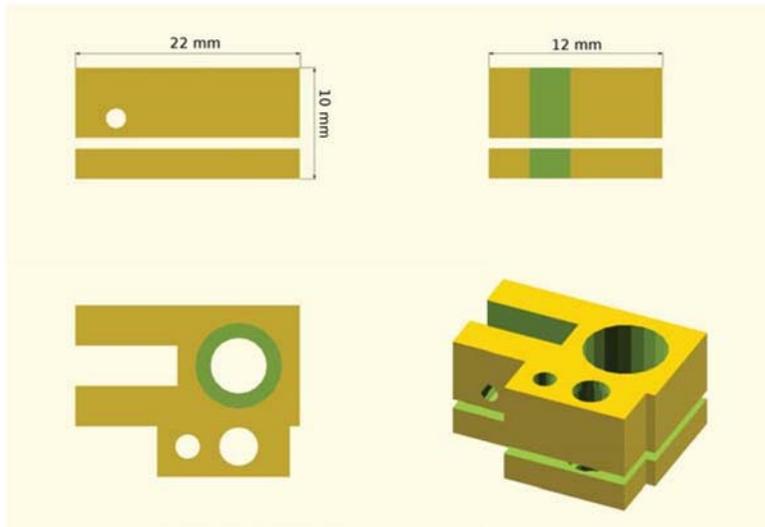

**3DPP15** component for actuating the syringe tips up and down.

### S.6.4.6  3DPP16, crank shaft

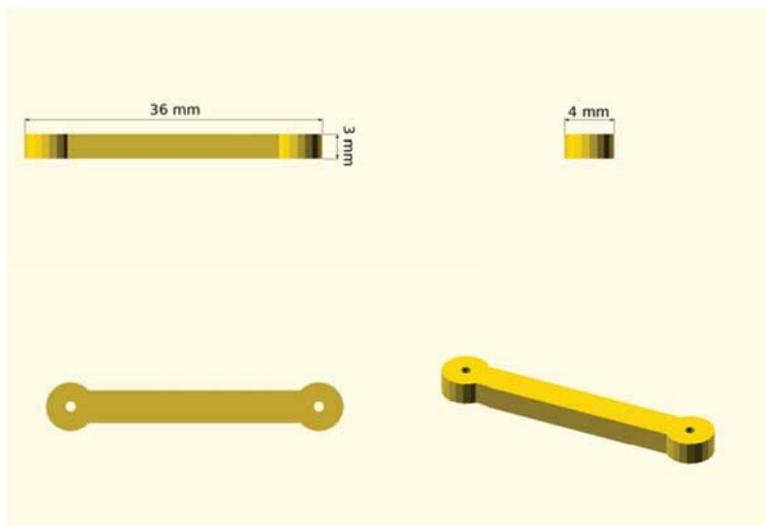

**3DPP16** crank shaft for syringe actuation, to attach to injection moulded motor arm.